\pdfoutput=1

\documentclass[dvipsnames,11pt]{article}

\usepackage[]{acl}

\usepackage{colortbl,xcolor,xspace}
\usepackage{multirow}
\usepackage{times}
\usepackage{hyperref}
\usepackage{latexsym}
\usepackage{circledsteps}
\usepackage{graphicx}
\usepackage{tikz}
\usepackage[T1]{fontenc}

\usepackage[algo2e]{algorithm2e} 
\usepackage{algorithm, algorithmic}
\usepackage{booktabs}
\usepackage{arydshln}
\usepackage{amsmath}
\usepackage{subcaption}
\usepackage[framemethod=TikZ]{mdframed}

\usepackage{listings}
\usepackage{spverbatim}
\usepackage{newfloat}
\usepackage{listings}

\usepackage{booktabs}

\usepackage[T1]{fontenc}    
\newcommand\funfont[1]{{\usefont{T1}{QTWestEndRegular}{m}{n}#1}}
\newcommand{\dataname}{\funfont{Compun}\xspace}

\usepackage[utf8]{inputenc}

\usepackage{microtype}

\usepackage[utf8]{inputenc}

\usepackage{microtype}

\usepackage{inconsolata}

\usepackage{fdsymbol}
\newcommand\blfootnote[1]{%
  \begingroup
  \renewcommand\thefootnote{}\footnote{#1}%
  \addtocounter{footnote}{-1}%
  \endgroup
}
\title{Do Vision-Language Models Understand Compound Nouns?}

\author{Sonal Kumar$^{\spadesuit*}$ \quad Sreyan Ghosh$^{\spadesuit*}$ \quad
\bf S Sakshi$^{\spadesuit}$ \quad Utkarsh Tyagi$^{\spadesuit}$ \\
\bf Dinesh Manocha$^{\spadesuit}$  \\
        $^{\spadesuit}$University of Maryland, College Park, USA \\
         \texttt{\{sonalkum, sreyang, utkarsht, dmanocha\}@umd.edu}}

\begin{document}
\maketitle
\begin{abstract}
Open-vocabulary vision-language models (VLMs) like CLIP, trained using contrastive loss, have emerged as a promising new paradigm for text-to-image retrieval. However, do VLMs understand compound nouns (CNs) (e.g., \textit{lab coat}) as well as they understand nouns (e.g., \textit{lab})? We curate \dataname, a novel benchmark with 400 unique and commonly used CNs, to evaluate the effectiveness of VLMs in interpreting CNs. The \dataname benchmark challenges a VLM for text-to-image retrieval where, given a text prompt with a CN, the task is to select the correct image that shows the CN among a pair of distractor images that show the constituent nouns that make up the CN. Next, we perform an in-depth analysis to highlight CLIPs' limited understanding of certain types of CNs. Finally, we present an alternative framework that moves beyond hand-written templates for text prompts widely used by CLIP-like models. We employ a Large Language Model to generate multiple diverse captions that include the CN as an object in the scene described by the caption. Our proposed method improves CN understanding of CLIP by 8.25\% on \dataname. Code and benchmark are available~\footnote{https://github.com/sonalkum/Compun}.\blfootnote{$^*$These authors contributed equally to this work.}

\end{abstract}


\section{Introduction}
\label{sec:intro}
A compound noun (CN) is a noun formed from two or more words combined to create a single noun with a new meaning. A CN usually combines two nouns (noun + noun, e.g., \textit{paper towel}) or an adjective and a noun (adjective + noun, e.g., \textit{full moon}); however, there exist more types, and we show an exhaustive list with examples in Appendix~\ref{subsec:types}. For the scope of this paper, we focus primarily on the noun + noun type.

Interpreting the meaning of CNs by decoding the implicit semantic relation between their constituent nouns has attracted great interest in Natural Language Processing (NLP) for decades~\cite{wisniewski1997concepts,coil2023chocolate}. This task requires systems to move beyond memorization as CNs are continually emerging, with new combinations frequently appearing~\cite{pinter-etal-2020-will}. Pre-trained Language Models (PLMs) that are trained on vast amounts of text and acquire broad semantic knowledge in the process have shown impressive performance in interpreting CNs, including unseen CNs~\cite{coil2023chocolate}. The improvements can also be partly attributed to the transformer architecture, which by design computes a word representation as a function of the representation of its context~\cite{coil2023chocolate}.

Though extensively studied in NLP, whether modern vision-language models (VLMs) understand CNs is under-explored. Open-vocabulary VLMs~\footnote{Our work only investigates VLMs trained with contrastive loss as they are widely adopted for retrieval tasks~\cite{ray2023cola}. Investigating other kinds of VLMs (e.g., autoregressive) is beyond the scope of our work.} like CLIP~\cite{radford2021learning}, trained using a contrastive loss between image-caption pairs, have become the go-to choice for image-to-text (zero-shot classification) and text-to-image retrieval~\cite{ray2023cola}. However, recent work shows that CLIP-like VLM models often act as bag of words and lack understanding of relationships between objects and attributes~\cite{yuksekgonul2023when}. For example, prior works explore the failure of VLMs to understand spatial relationships between two objects in the image through the caption (e.g., ``left of'')~\cite{kamath2023whatsup} or the binding of a verb with its corresponding object (e.g., ``running tiger''). To the best of our knowledge, evaluating a VLM's understanding of the semantic relationship between nouns to interpret CNs hasn't been explored in literature.
\vspace{1mm}

%


{\noindent \textbf{Main Contributions.}} We propose \dataname, a benchmark with 400 instances that serves as a test bed to evaluate a VLM's ability to interpret CNs. Each instance in \dataname corresponds to a unique compound noun and includes one image representing the compound noun, along with two additional distractor images. These distractor images depict the individual constituent nouns that form the CN (investigating CNs with more than 2 nouns remains part of future work.) (example in Fig.~\ref{fig:main_diag}). Given the class name (or the CN), the task of a VLM is to retrieve (or select) the correct image among the distractors. We perform a detailed analysis of CLIPs' performance on \dataname, providing an in-depth understanding of how well state-of-the-art VLMs interpret CNs. Next, we present a novel framework to improve text-to-image retrieval that moves beyond generic hand-written prompts for text-to-image retrieval. Given a CN, we generate multiple diverse captions using an LLM, where each caption describes a scene with the CN as a key object in it. Finally, the captions are used to construct a custom prompt for text-to-image retrieval. Our proposed method improves CLIP's and OpenCLIP's performance by 8.25\% and 2.35\%, respectively, on \dataname.


\section{Background and Related Work}
\label{sec:related_work}

{\noindent \textbf{Interpreting CNs.}} Cognitive science research has examined human processing of novel noun-noun pairings~\cite{wisniewski1997concepts,costello2000efficient,connell2012flexible}. Although these pairings can lead to multiple interpretations, typically, one interpretation emerges as the most naturally comprehensible~\cite{costello2000efficient}. Early work in interpreting compound nouns has majorly framed the task as a classification task, where each compound noun is classified to a single relation~\cite{kim2005automatic,tratz2010taxonomy}. However, owing to the ambiguity where a single compound noun can be classified into multiple relations~\cite{shwartz-dagan-2018-paraphrase}, recent work has adopted paraphrasing for the same task~\cite{kim-nakov-2011-large,pasca2015interpreting,shwartz-dagan-2018-paraphrase}. The task is again to classify a compound noun into multiple pre-defined templates. \citet{ponkiya-etal-2020-looking} show that PLMs acquire adequate knowledge to understand the semantic relationship between the constituent nouns in a CN during pre-training itself. Following this, a wealth of work employs sequence-to-sequence PLMs (including LLMs) to assess their ability to interpret existing and novel blends of nouns~\cite{shwartz2021long,li-etal-2022-systematicity,pinter-etal-2020-will}.
\vspace{0.5mm}

{\noindent \textbf{Contrastive VLMs.}} Contrastive VLMs include models trained using a contrastive loss between image-caption pairs. CLIP~\cite{radford2019language}, a pioneering work in this space, shows that such a model can improve text-to-image and image-to-text retrieval, with applications in zero-shot classification, etc. Following CLIP, a wealth of work focuses on improving different aspects of CLIP, like compositionality~\cite{nayak2023learning}, and also employ CLIP as a backbone vision encoder for various vision tasks like captioning~\cite{mokady2021clipcap}, instruction following~\cite{liu2023visual}, etc. Our work is inspired by the fact that contrastive VLMs often act as bag of words~\cite{yuksekgonul2023when} and might lack an understanding of the semantic relationship between the constituent nouns to interpret the final CN.


\section{\dataname Benchmark}
\label{sec:compun}

{\noindent \textbf{The task.}} Our \dataname benchmark serves as a test bed for evaluating a VLM's capability to interpret compound nouns. For evaluation, we focus on the zero-shot text-to-image retrieval task, where given a natural language prompt, the task of a VLM is to retrieve an image that illustrates the image described in the prompt. In the base setting, our prompt just describes a compound noun as \textit{``A photo of a \{compound noun\}''}. Text-to-image retrieval has been earlier adopted by several works for evaluating compositional understanding~\cite{yuksekgonul2023when,ray2023cola}. Inspired by these works, we design the Compun benchmark to challenge a VLM to select the correct image among a set of distractors. More precisely, each instance in the \dataname benchmark, attributed to a compound noun, has 3 images, where only one image illustrates the compound noun, while the other two images illustrate the constituent nouns that make up the compound noun (example in Figure~\ref{fig:main_diag}). All compound nouns in the \dataname benchmark have a maximum of two nouns. 
\vspace{0.5mm}

{\noindent \textbf{Evaluation.}} We resort to a simple evaluation metric, consistent with prior-art~\cite{thrush2022winoground} for evaluating a VLM on \dataname. Formally, let us denote the image illustrating the compound noun as a positive ($\mathcal{P}$) and the other 2 distractor images as negatives ($\mathcal{N}_1$ and $\mathcal{N}_2$). Thus, given the natural language prompt $\mathcal{C}$ for the compound noun , our evaluation metric $f\left(\mathcal{C}, \mathcal{P}, \mathcal{N}_1, \mathcal{N}_2\right)$ is defined as:
\vspace{-1mm}
\begin{equation}
\label{eqtn:eval}
\small
f\left(\mathcal{C}, \mathcal{P}, \mathcal{N}_1, \mathcal{N}_2\right)= \begin{cases}1 & \text { if } s\left(\mathcal{C}, \mathcal{P}\right)>s\left(\mathcal{C}, \mathcal{N}_1\right) \\ & \text { and } s\left(\mathcal{C}, \mathcal{P}\right)>s\left(\mathcal{C}, \mathcal{N}_2\right)\\ 0 & \text { otherwise }\end{cases}
\end{equation}

where $s(.)$ is the standard cosine similarity, widely used for retrieval. 

\vspace{0.5mm}

{\noindent \textbf{Data collection and annotation.}} The \dataname benchmark has 400 test instances and a total of 1200 (400$\times$3) images. Each instance in the \dataname benchmark is attributed to a unique compound noun (the complete list is provided in Table~\ref{subsec:list}). We use a combination of compound nouns provided by \citet{levin2019systematicity}, \citet{lang2022visually}, GPT-4~\cite{openai2023gpt4}, and the internet. Next, we discard compound nouns that can have confusing distractors (e.g., \textit{cheesecake}, where it's usually hard to distinguish between a \textit{cheesecake} and any other \textit{cake}). After this step, we filter 400 compound nouns, the most widely used from our list. While a compound noun can have multiple interpretations, we use the more commonly known one. For a compound noun that may have multiple interpretations, we use MTurk to decide the most commonly known one. More details about this study can be found in Appendix~\ref{subsec:mturk}. Finally, a group of 4 annotators collects the required 1200 images from various image search engines. All 4 annotators come with extensive vision and language research experience.

\section{Retrieval with Example Captions}
\label{sec:diverse}

Fig.~\ref{fig:main_diag} illustrates our proposed approach. As discussed earlier, the standard approach for text-to-image retrieval using class names is to hand-write several prompt templates (e.g., ``a photo of a {class name}.''). We propose an alternative framework -- retrieve with example captions. Our framework is zero-shot and requires no further training. Given a compound noun $c$, we ask an LLM to generate 5 diverse captions, where each caption has the compound noun $c$ as an object in it. The generated captions should have $c$ in diverse settings with diverse adjectives and verbs. We instruct GPT-4~\cite{openai2023gpt4} with the following prompt to generate the captions:

\begin{figure}
    \centering
    \includegraphics[width=\columnwidth]{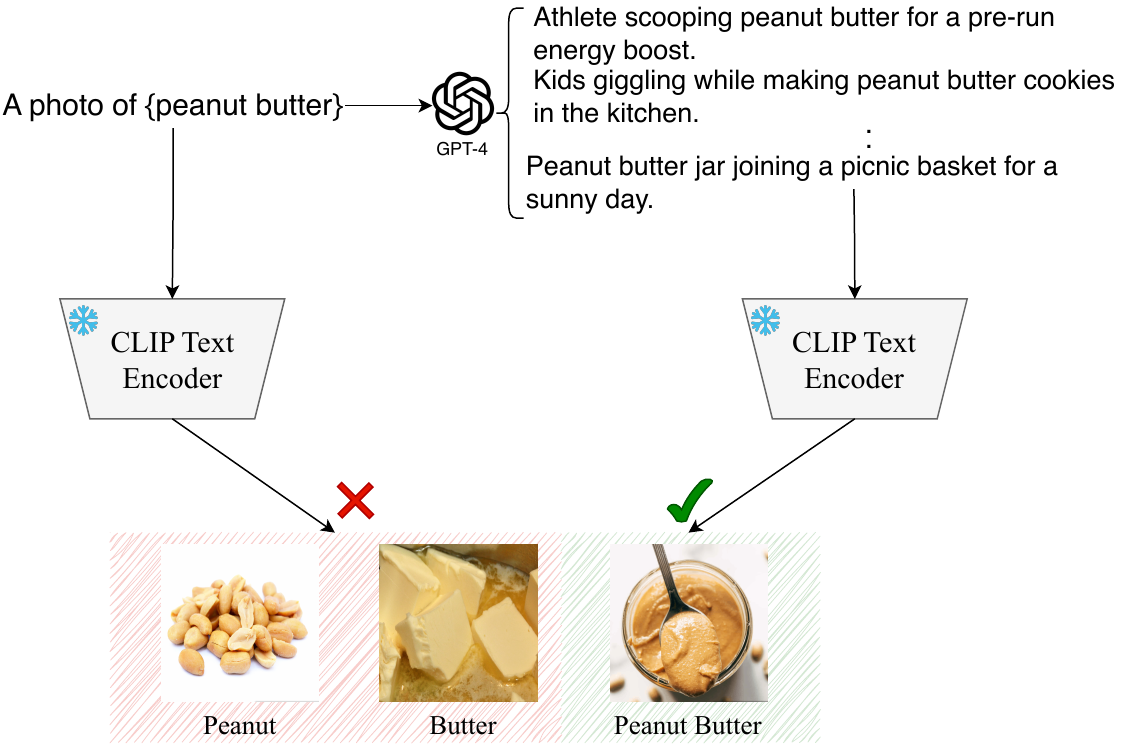}
    \caption{\small Illustration of our proposed \textbf{Retrieval with Captions}. We first generate 5 diverse captions describing 5 diverse scenes, with the compound noun as an object in it. These captions are then used to build 5 custom text prompts for text-to-image retrieval, and the image with the highest mean similarity to all 5 prompts is then selected for retrieval.}
    \label{fig:main_diag}
    \vspace{-1em}
\end{figure}

\begin{mdframed}[backgroundcolor=gray!20, linecolor=white, innerleftmargin=10pt, innerrightmargin=10pt, innertopmargin=10pt, innerbottommargin=10pt]
Return a list of 5  diverse captions with a compound\_noun in a photo. The captions should be a maximum of 10 words and one-liners. All 5 captions should describe the compound noun in diverse settings with different verbs and actions being performed with the compound noun. An example output for "chicken burger": ['Sizzling chicken burger grilling at a lively backyard BBQ.,' 'Chef expertly flipping a juicy chicken burger in a diner.',' Family enjoying homemade chicken burgers on a sunny picnic.', 'Athlete fueling up with a protein-packed chicken burger post-workout.', 'Friends sharing a chicken burger at a vibrant street festival.']. Only return a list of strings and nothing else.
\end{mdframed}

and an example output for the CN \textit{"chocolate crocodile"} is as follows:
\begin{figure*}[t!]
    \centering
    \includegraphics[width=2\columnwidth]{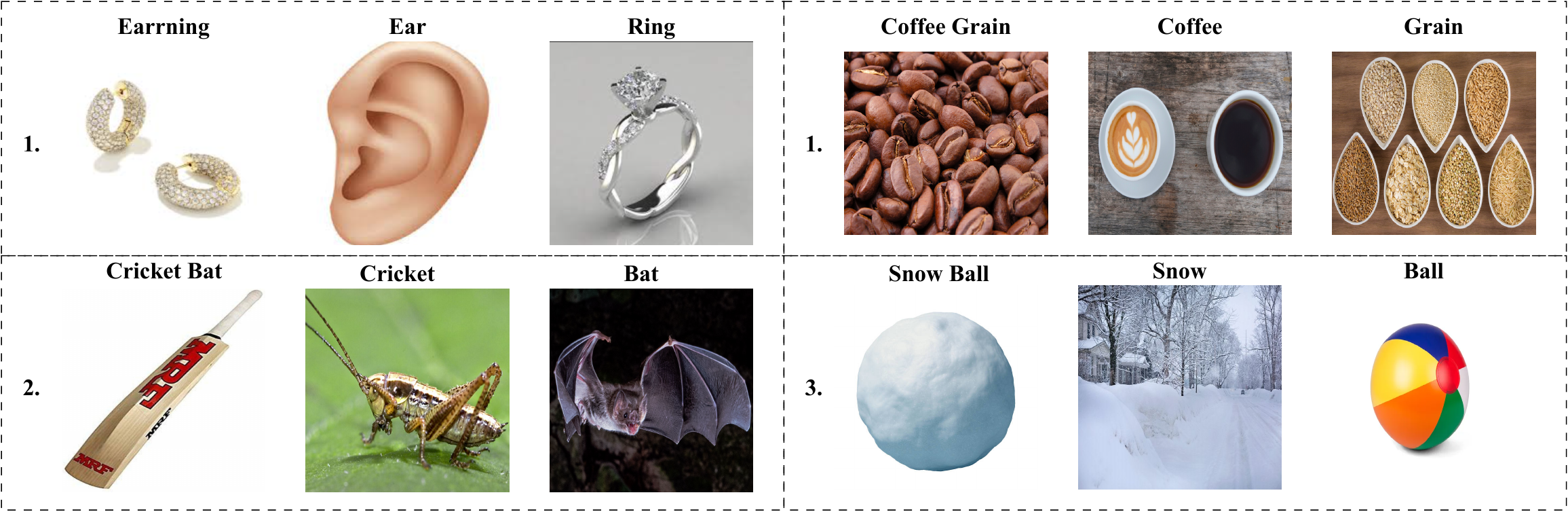}
    \caption{\small Illustration of 3 types of CNs used in our study: Either Noun, Both Nouns and None. A brief explanation of the 3 types is provided in Section~\ref{sec:analysis}. \textbf{1. (left)} An example of Either Noun, where \textit{earring} looks like an ordinary \textit{ring} but not like an \textit{ear}, and the noun \textit{ear} just acts as an attribute that modifies the visual of a \textit{ring} to an \textit{earring}. \textbf{1. (right)} An example of  Either Noun, where \textit{coffee grain} looks like an ordinary grain but is modified by the noun \textit{coffee}, which acts as an attribute. \textbf{2.} An example of None, where a \textit{cricket bat} looks completely different from both \textit{cricket} and \textit{bat}. \textbf{3.} An example of Both Nouns, where a \textit{snow ball} looks both like \textit{snow} and \textit{ball}.}
    \label{fig:types_compun}
\end{figure*}

\begin{mdframed}[backgroundcolor=gray!20, linecolor=white, innerleftmargin=10pt, innerrightmargin=10pt, innertopmargin=10pt, innerbottommargin=10pt]
["Pastry chef sculpting a chocolate crocodile with finesse.", "Kids discovering a chocolate crocodile in a candy treasure hunt.", "Artist painting a whimsical chocolate crocodile in a foodie gallery.", "Chocolate crocodile starring in a whimsical patisserie window display.", "Chocolate crocodile sunbathing on a dessert island paradise."]
\end{mdframed}

We then build a prompt for our VLM separately with each of the captions as follows to get 5 final prompts: \textit{``a photo of a \{class name\}. An example of \{compound name\} in an image is \{caption\}''}. Next, we calculate the mean similarity of an image $c \in C$ with the text prompts as follows:
\vspace{-1.5mm}
\begin{equation}
    \text{Mean Similarity} = \frac{1}{n} \sum_{i=1}^{5} s(c, p_i)
\end{equation}

where $p_i \in P$ denotes the generated prompts, $s(.)$ is the standard cosine similarity formulation, and $c \in C$ denotes the set of all images available for text-to-image retrieval or every image the text has to be compared with. Finally, we choose the image with the highest mean cosine similarity.

Our core hypothesis builds on existing work in using language as an internal representation for visual recognition, which creates an interpretable bottleneck for computer vision tasks~\cite{menon2022visual,pratt2023does}. Instead of querying a VLM with just the compound noun, employing language enables us to compare to any words flexibly. Since interpreting compound nouns is easier when provided with proper context in example sentences, getting exposed to diverse keywords through examples makes the image with the compound noun a strongly activating image while the distractors are lowly activating. Taking an example from Fig.~\ref{fig:main_diag}, keywords like \textit{player} and \textit{wooden} obtained through diverse captions make the original image more activating than its distractors. Our proposed method also improves performance on benchmark text-to-image retrieval datasets, and we provide additional results in Appendix~\ref{sec:imagenet}.



\section{Experiments and Results}
\label{sec:results}

{\noindent \textbf{Baselines.}} For our baselines, we employ the original CLIP~\cite{radford2019language}, OpenCLIP~\cite{ilharco_gabriel_2021_5143773}, ALIGN~\cite{jia2021scaling}, ALBEF~\cite{li2021align}, BLIP~\cite{li2022blip} and MetaCLIP~\cite{Zhai_2023_ICCV}. All these methods are trained with contrastive learning on image-text pairs. We also employ CLIP~\textit{w/ desc}~\cite{menon2022visual}, which adds image descriptors to the prompt for retrieval. Finally, we also ablate with CLIP \textit{rev.} where we switch the order of nouns in the compound noun. We prompt GPT-4 with a temperature of 0.1 and top-p of 1.
\vspace{0.5mm}

{\noindent \textbf{Results.}} Table~\ref{tab:results} compares the performance of various VLMs on the \dataname benchmark. We also perform a human evaluation on our benchmark using MTurk. While OpenCLIP achieves the highest performance with simple template prompts, our method improves OpenCLIP performance by 2.35\%. Similarly, our method improves CLIP performance by 8.24\%. CLIP~\textit{rev.} leads to a 37.25\% drop in performance over CLIP, indicating that CLIP has some understanding of the semantic relationship between the nouns. In the next section, we make important conclusions regarding CLIP's limited understanding of \textit{attributed compound nouns}.

\begin{table}[h]
\centering
\resizebox{0.9\columnwidth}{!}{%
\begin{tabular}{lc}
\toprule
\textbf{Model}    & \textbf{Text-to-Image Acc.}  \\
\midrule
Human    &  96.25\\
Random     &  33.33\\
\midrule
ALBEF~\cite{li2021align}    &  80.55\\
BLIP~\cite{li2022blip}    &  79.85\\  
MetaCLIP~\cite{xu2023metaclip} & 81.35\\
CLIP~\cite{radford2019language}     &  78.25 \\
CLIP~\textit{rev.}   &  41.00 \\
CLIP~\textit{w/ desc}~\cite{menon2022visual} &  81.15\\
\cellcolor{blue!10}CLIP~\textit{w/ examples (ours)}     &  \cellcolor{blue!10}86.50\\
OpenCLIP~\cite{ilharco_gabriel_2021_5143773} &   83.90 \\
\cellcolor{blue!10}OpenCLIP~\textit{w/ examples (ours)} &   \cellcolor{blue!10}86.25 \\ \bottomrule
\end{tabular}%
}
\caption{\small Comparison of our proposed version of CLIP with other baselines on the \dataname benchmark. Our proposed method outperforms CLIP by 8.25\% and OpenCLIP by 2.35\%.}
\label{tab:results}
\vspace{-1em}
\end{table}

\begin{figure}[t]
    \centering
    \includegraphics[width=0.8\columnwidth]{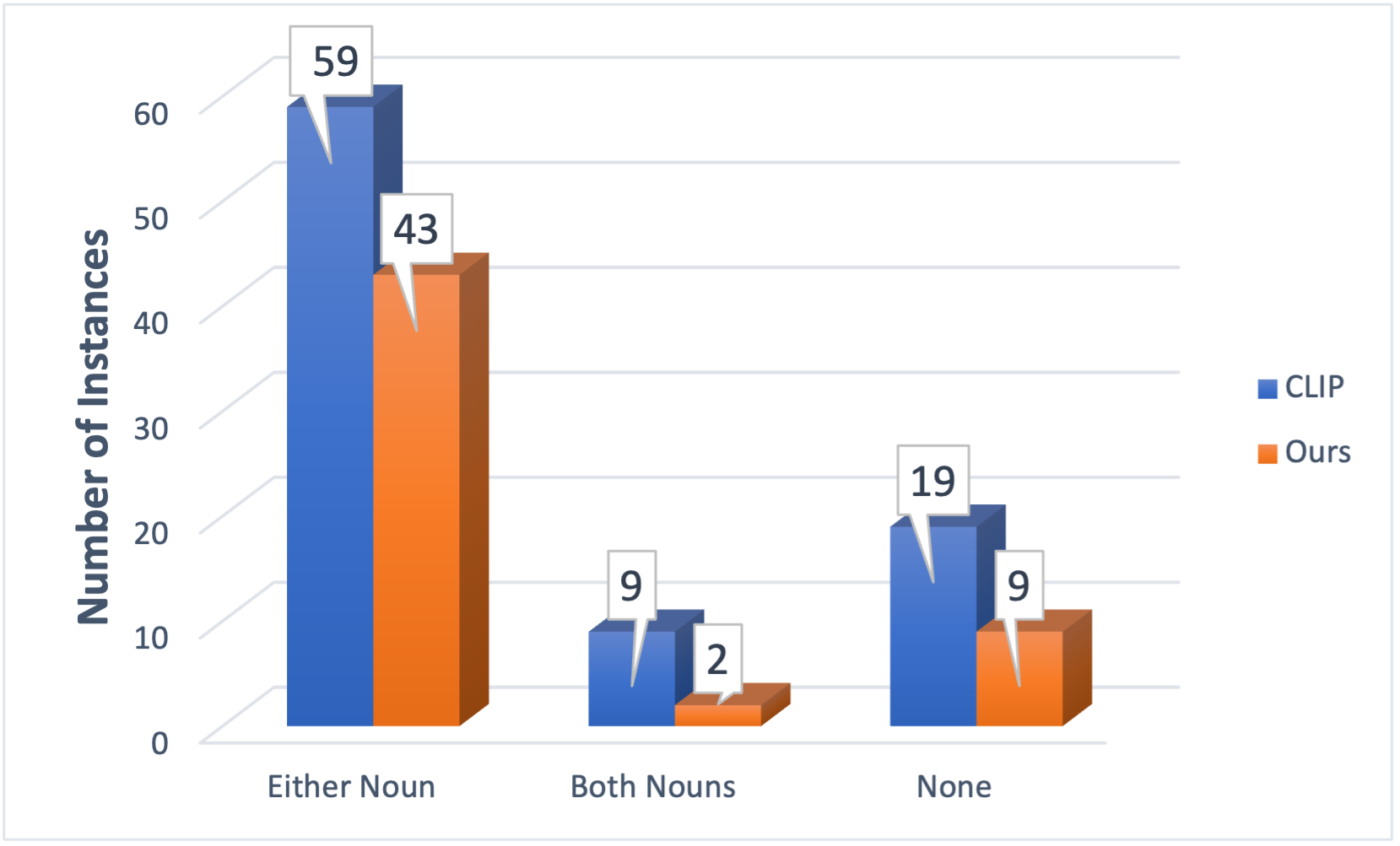}
    \caption{\small Count of misclassified instances by CLIP on \dataname for three settings, either, both, and none. Section~\ref{sec:analysis} describes these settings. CLIP is more likely to retrieve a negative when the positive image shows either constituent noun, highlighting CLIP's limited understanding of attributed CNs.}

        \label{fig:noun_chart}
\end{figure}
\begin{figure}[h]
    \centering
    \includegraphics[width=0.8\columnwidth]{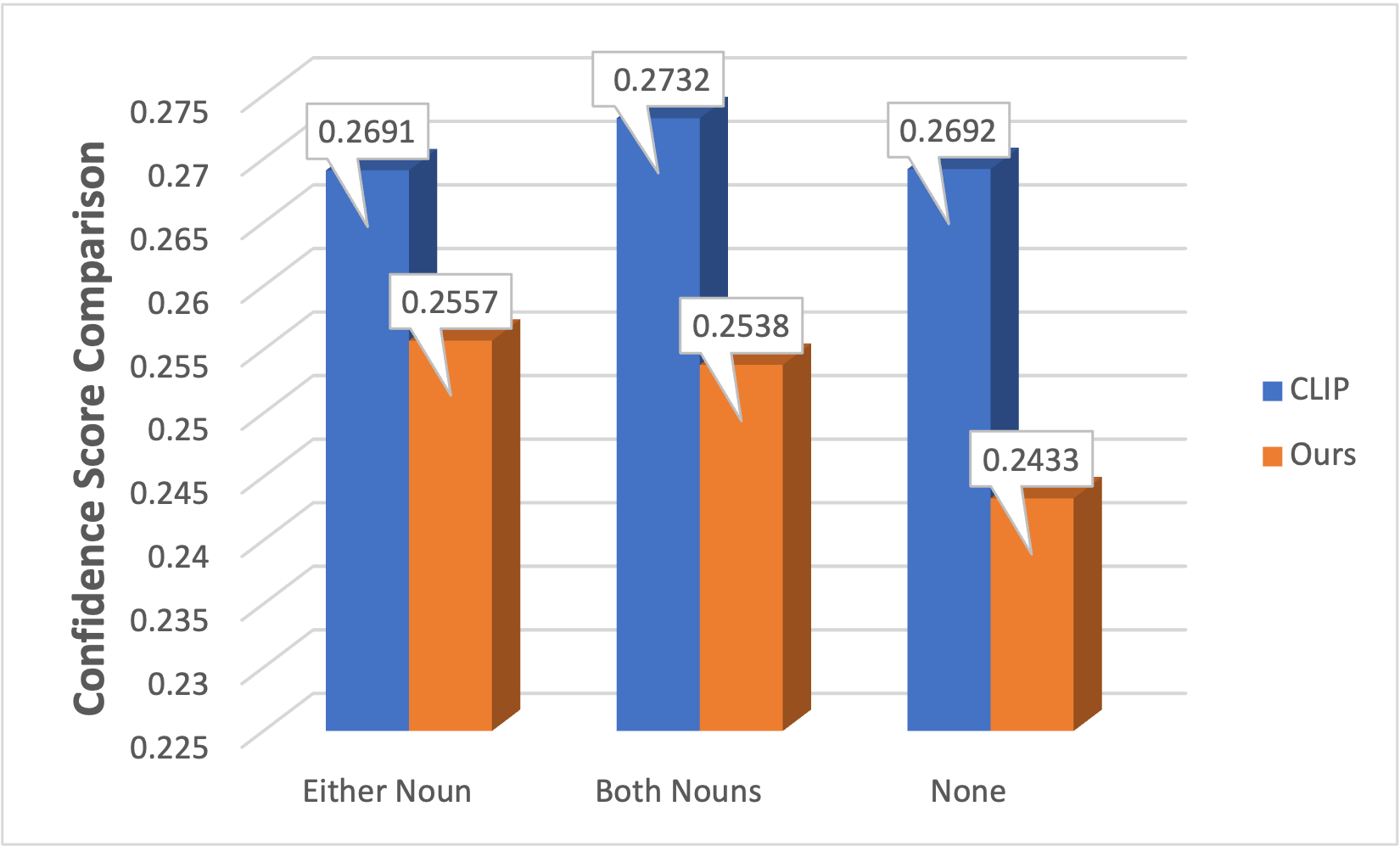}
    \caption{\small Average CLIP similarity scores for correct predictions on \dataname on three unique settings, either, both, and none. Section~\ref{sec:analysis} describes these settings. High scores on the \dataname benchmark are superficial, and CLIP often wins by low similarity scores.}
    \label{fig:conf_chart}
    \vspace{-1em}
\end{figure}

\section{Results Analysis}
\label{sec:analysis}

To perform an in-depth analysis of the results, we first perform an MTurk study to divide all CNs in \dataname into 3 main categories as illustrated in Fig.~\ref{fig:types_compun}: \textbf{1.}CNs that clearly show either of their constituent nouns in the picture. In this case, one noun acts as an attribute to the other, changing its visual minimally, but is not visible itself (e.g., \textit{coffee grain}). \textbf{2.}CNs that clearly show both the constituent nouns in the picture. This is the same as \textbf{1.}, but both nouns are visible (e.g., \textit{snow ball}) and \textbf{3.} CNs that show none of the constituent nouns in the picture and make up a completely new CN (e.g., \textit{cricket bat}). The 3 settings have 199, 106, and 95 instances in \dataname. Fig.~\ref{fig:noun_chart} compares the number of incorrectly predicted instances in \dataname across these 3 categories. CLIP makes the highest number of mistakes in the first category, which also indicates CLIPs' limited understanding of such CNs, which can also be interpreted as attributed CNs. Such CNs are also emerging in nature~\cite{coil2023chocolate}, and correctly interpreting them is a long-standing problem in NLP. On the other hand, CLIP makes the least mistakes in the third type, indicating that CLIP might have acquired adequate knowledge about unique objects in its pre-training stage.  Fig.~\ref{fig:conf_chart} shows that results on the \dataname benchmark are superficially low -- similarity scores for correct predictions are $\approx$25\textsubscript{+-2\%} (in contrast to ImageNet retrieval with $\approx$82\%). We perform retrieval, treating the entire benchmark images as negatives, and achieve a score of 12\%, a 66.25\% drop. 







\section{Conclusion}
\label{sec:conclusion}
This paper presents the first study of VLMs in interpreting CNs. We curate \dataname, a novel benchmark with 400 unique CNs, and show that CLIP has a limited understanding of CNs where one of the two constituent nouns acts as an attribute to the other. Next, we present a novel approach that moves beyond generic template-based prompts and leverages an LLM to generate diverse captions describing scenes with the CN as an object in the scene. Our proposed method improves the performance of CLIP on \dataname significantly.

\section*{Limitations and Future Work}
\label{sec:limitations}

We list down some potential limitations of our work:

\begin{enumerate}
    \item \dataname focuses on this sole definition of CN interpretation -- Can VLMs distinguish between a CN and its constituent nouns? \dataname does not consist of emerging CNs like the NYTWIT dataset~\cite{pinter-etal-2020-nytwit}. This dataset proposed CNs where humans created entirely new CNs using editing nouns corresponding to entirely new concepts. These CNs are particularly challenging for even modern LLMs to interpret and require strong reasoning abilities over context~\cite{coil2023chocolate}. However, after a preliminary analysis, we hypothesize that most of the CNs in ~\citet{pinter-etal-2020-nytwit} are rare, and VLMs might not have come across these CNs or concepts from their pre-training stage. We want to explore this as part of future work as this brings entirely new challenges to VLMs, including complex reasoning abilities.
    \item \dataname, like other text-to-image retrieval benchmarks, would benefit from better evaluation metrics. Though our metrics are inspired by prior art, as shown in Section~\ref{sec:analysis}, results on \dataname are superficial, and VLMs can perform well even with low confidence scores (corresponding to low activations when the VLM sees the CN). Additionally, evaluating \dataname with the entire benchmark as negatives makes it difficult to gain an understanding of where and how VLMs go wrong in interpreting CNs. Thus, as part of future work, we would like to explore better evaluation metrics and benchmark design.

    \item We evaluate \dataname only on contrastive trained VLMs as we try to study CN interpretation through the lens of text-to-image retrieval, and contrastive VLMs fit well to the retrieval task. As part of future work, we would like to study how well other types of VLMs, like auto-regressive~\cite{liu2023visual} VLMs, interpret CNs.
    
\end{enumerate}


\bibliography{anthology,custom}

\begin{thebibliography}{34}
\expandafter\ifx\csname natexlab\endcsname\relax\def\natexlab#1{#1}\fi

\bibitem[{Coil and Shwartz(2023)}]{coil2023chocolate}
Jordan Coil and Vered Shwartz. 2023.
\newblock From chocolate bunny to chocolate crocodile: Do language models understand noun compounds?
\newblock \emph{arXiv preprint arXiv:2305.10568}.

\bibitem[{Connell and Lynott(2012)}]{connell2012flexible}
Louise Connell and Dermot Lynott. 2012.
\newblock Flexible shortcuts: Linguistic distributional information affects both shallow and deep conceptual processing.
\newblock In \emph{Proceedings of the Annual Meeting of the Cognitive Science Society}, volume~34.

\bibitem[{Costello and Keane(2000)}]{costello2000efficient}
Fintan~J Costello and Mark~T Keane. 2000.
\newblock Efficient creativity: Constraint-guided conceptual combination.
\newblock \emph{Cognitive Science}, 24(2):299--349.

\bibitem[{Ilharco et~al.(2021)Ilharco, Wortsman, Wightman, Gordon, Carlini, Taori, Dave, Shankar, Namkoong, Miller, Hajishirzi, Farhadi, and Schmidt}]{ilharco_gabriel_2021_5143773}
Gabriel Ilharco, Mitchell Wortsman, Ross Wightman, Cade Gordon, Nicholas Carlini, Rohan Taori, Achal Dave, Vaishaal Shankar, Hongseok Namkoong, John Miller, Hannaneh Hajishirzi, Ali Farhadi, and Ludwig Schmidt. 2021.
\newblock \href {https://doi.org/10.5281/zenodo.5143773} {Openclip}.
\newblock If you use this software, please cite it as below.

\bibitem[{Jia et~al.(2021)Jia, Yang, Xia, Chen, Parekh, Pham, Le, Sung, Li, and Duerig}]{jia2021scaling}
Chao Jia, Yinfei Yang, Ye~Xia, Yi-Ting Chen, Zarana Parekh, Hieu Pham, Quoc Le, Yun-Hsuan Sung, Zhen Li, and Tom Duerig. 2021.
\newblock Scaling up visual and vision-language representation learning with noisy text supervision.
\newblock In \emph{International conference on machine learning}, pages 4904--4916. PMLR.

\bibitem[{Kamath et~al.(2023)Kamath, Hessel, and Chang}]{kamath2023whatsup}
Amita Kamath, Jack Hessel, and Kai-Wei Chang. 2023.
\newblock "what's 'up' with vision-language models? investigating their struggle to understand spatial relations.".
\newblock In \emph{EMNLP}.

\bibitem[{Kim and Baldwin(2005)}]{kim2005automatic}
Su~Nam Kim and Timothy Baldwin. 2005.
\newblock Automatic interpretation of noun compounds using wordnet similarity.
\newblock In \emph{International Conference on Natural Language Processing}, pages 945--956. Springer.

\bibitem[{Kim and Nakov(2011)}]{kim-nakov-2011-large}
Su~Nam Kim and Preslav Nakov. 2011.
\newblock \href {https://aclanthology.org/D11-1060} {Large-scale noun compound interpretation using bootstrapping and the web as a corpus}.
\newblock In \emph{Proceedings of the 2011 Conference on Empirical Methods in Natural Language Processing}, pages 648--658, Edinburgh, Scotland, UK. Association for Computational Linguistics.

\bibitem[{Lang et~al.(2022)Lang, van~der Plas, Nissim, Gatt et~al.}]{lang2022visually}
Inga Lang, Lonneke van~der Plas, Malvina Nissim, Albert Gatt, et~al. 2022.
\newblock Visually grounded interpretation of noun-noun compounds in english.
\newblock In \emph{Proceedings of the Workshop on Cognitive Modelling and Computational Linguistics (CMCL'22)}. Association for Computational Linguistics.

\bibitem[{Levin et~al.(2019)Levin, Glass, and Jurafsky}]{levin2019systematicity}
Beth Levin, Lelia Glass, and Dan Jurafsky. 2019.
\newblock Systematicity in the semantics of noun compounds: The role of artifacts vs. natural kinds.
\newblock \emph{Linguistics}, 57(3):429--471.

\bibitem[{Li et~al.(2022{\natexlab{a}})Li, Li, Xiong, and Hoi}]{li2022blip}
Junnan Li, Dongxu Li, Caiming Xiong, and Steven Hoi. 2022{\natexlab{a}}.
\newblock Blip: Bootstrapping language-image pre-training for unified vision-language understanding and generation.
\newblock In \emph{International Conference on Machine Learning}, pages 12888--12900. PMLR.

\bibitem[{Li et~al.(2021)Li, Selvaraju, Gotmare, Joty, Xiong, and Hoi}]{li2021align}
Junnan Li, Ramprasaath Selvaraju, Akhilesh Gotmare, Shafiq Joty, Caiming Xiong, and Steven Chu~Hong Hoi. 2021.
\newblock Align before fuse: Vision and language representation learning with momentum distillation.
\newblock \emph{Advances in neural information processing systems}, 34:9694--9705.

\bibitem[{Li et~al.(2022{\natexlab{b}})Li, Carlson, and Potts}]{li-etal-2022-systematicity}
Siyan Li, Riley Carlson, and Christopher Potts. 2022{\natexlab{b}}.
\newblock \href {https://doi.org/10.18653/v1/2022.findings-emnlp.50} {Systematicity in {GPT}-3{'}s interpretation of novel {E}nglish noun compounds}.
\newblock In \emph{Findings of the Association for Computational Linguistics: EMNLP 2022}, pages 717--728, Abu Dhabi, United Arab Emirates. Association for Computational Linguistics.

\bibitem[{Liu et~al.(2023)Liu, Li, Wu, and Lee}]{liu2023visual}
Haotian Liu, Chunyuan Li, Qingyang Wu, and Yong~Jae Lee. 2023.
\newblock Visual instruction tuning.
\newblock \emph{arXiv preprint arXiv:2304.08485}.

\bibitem[{Menon and Vondrick(2023)}]{menon2022visual}
Sachit Menon and Carl Vondrick. 2023.
\newblock Visual classification via description from large language models.
\newblock \emph{ICLR}.

\bibitem[{Mokady et~al.(2021)Mokady, Hertz, and Bermano}]{mokady2021clipcap}
Ron Mokady, Amir Hertz, and Amit~H Bermano. 2021.
\newblock Clipcap: Clip prefix for image captioning.
\newblock \emph{arXiv preprint arXiv:2111.09734}.

\bibitem[{Nayak et~al.(2023)Nayak, Yu, and Bach}]{nayak2023learning}
Nihal~V. Nayak, Peilin Yu, and Stephen Bach. 2023.
\newblock \href {https://openreview.net/forum?id=S8-A2FXnIh} {Learning to compose soft prompts for compositional zero-shot learning}.
\newblock In \emph{The Eleventh International Conference on Learning Representations}.

\bibitem[{OpenAI(2023)}]{openai2023gpt4}
OpenAI. 2023.
\newblock \href {http://arxiv.org/abs/2303.08774} {Gpt-4 technical report}.

\bibitem[{Pasca(2015)}]{pasca2015interpreting}
Marius Pasca. 2015.
\newblock Interpreting compound noun phrases using web search queries.
\newblock In \emph{Proceedings of the 2015 Conference of the North American Chapter of the Association for Computational Linguistics: Human Language Technologies}, pages 335--344.

\bibitem[{Pinter et~al.(2020{\natexlab{a}})Pinter, Jacobs, and Bittker}]{pinter-etal-2020-nytwit}
Yuval Pinter, Cassandra~L. Jacobs, and Max Bittker. 2020{\natexlab{a}}.
\newblock \href {https://doi.org/10.18653/v1/2020.coling-main.572} {{NYTWIT}: A dataset of novel words in the {N}ew {Y}ork {T}imes}.
\newblock In \emph{Proceedings of the 28th International Conference on Computational Linguistics}, pages 6509--6515, Barcelona, Spain (Online). International Committee on Computational Linguistics.

\bibitem[{Pinter et~al.(2020{\natexlab{b}})Pinter, Jacobs, and Eisenstein}]{pinter-etal-2020-will}
Yuval Pinter, Cassandra~L. Jacobs, and Jacob Eisenstein. 2020{\natexlab{b}}.
\newblock \href {https://doi.org/10.18653/v1/2020.findings-emnlp.138} {Will it unblend?}
\newblock In \emph{Findings of the Association for Computational Linguistics: EMNLP 2020}, pages 1525--1535, Online. Association for Computational Linguistics.

\bibitem[{Ponkiya et~al.(2020)Ponkiya, Murthy, Bhattacharyya, and Palshikar}]{ponkiya-etal-2020-looking}
Girishkumar Ponkiya, Rudra Murthy, Pushpak Bhattacharyya, and Girish Palshikar. 2020.
\newblock \href {https://doi.org/10.18653/v1/2020.findings-emnlp.386} {Looking inside noun compounds: Unsupervised prepositional and free paraphrasing}.
\newblock In \emph{Findings of the Association for Computational Linguistics: EMNLP 2020}, pages 4313--4323, Online. Association for Computational Linguistics.

\bibitem[{Pratt et~al.(2023)Pratt, Covert, Liu, and Farhadi}]{pratt2023does}
Sarah Pratt, Ian Covert, Rosanne Liu, and Ali Farhadi. 2023.
\newblock What does a platypus look like? generating customized prompts for zero-shot image classification.
\newblock In \emph{Proceedings of the IEEE/CVF International Conference on Computer Vision}, pages 15691--15701.

\bibitem[{Radford et~al.(2021)Radford, Kim, Hallacy, Ramesh, Goh, Agarwal, Sastry, Askell, Mishkin, Clark et~al.}]{radford2021learning}
Alec Radford, Jong~Wook Kim, Chris Hallacy, Aditya Ramesh, Gabriel Goh, Sandhini Agarwal, Girish Sastry, Amanda Askell, Pamela Mishkin, Jack Clark, et~al. 2021.
\newblock Learning transferable visual models from natural language supervision.
\newblock In \emph{International conference on machine learning}, pages 8748--8763. PMLR.

\bibitem[{Radford et~al.(2019)Radford, Wu, Child, Luan, Amodei, Sutskever et~al.}]{radford2019language}
Alec Radford, Jeffrey Wu, Rewon Child, David Luan, Dario Amodei, Ilya Sutskever, et~al. 2019.
\newblock Language models are unsupervised multitask learners.
\newblock \emph{OpenAI blog}, 1(8):9.

\bibitem[{Ray et~al.(2023)Ray, Radenovic, Dubey, Plummer, Krishna, and Saenko}]{ray2023cola}
Arijit Ray, Filip Radenovic, Abhimanyu Dubey, Bryan~A Plummer, Ranjay Krishna, and Kate Saenko. 2023.
\newblock Cola: A benchmark for compositional text-to-image retrieval.
\newblock In \emph{Thirty-seventh Conference on Neural Information Processing Systems Datasets and Benchmarks Track}.

\bibitem[{Shwartz(2021)}]{shwartz2021long}
Vered Shwartz. 2021.
\newblock A long hard look at mwes in the age of language models.
\newblock In \emph{Proceedings of the 17th Workshop on Multiword Expressions (MWE 2021)}.

\bibitem[{Shwartz and Dagan(2018)}]{shwartz-dagan-2018-paraphrase}
Vered Shwartz and Ido Dagan. 2018.
\newblock \href {https://doi.org/10.18653/v1/P18-1111} {Paraphrase to explicate: Revealing implicit noun-compound relations}.
\newblock In \emph{Proceedings of the 56th Annual Meeting of the Association for Computational Linguistics (Volume 1: Long Papers)}, pages 1200--1211, Melbourne, Australia. Association for Computational Linguistics.

\bibitem[{Thrush et~al.(2022)Thrush, Jiang, Bartolo, Singh, Williams, Kiela, and Ross}]{thrush2022winoground}
Tristan Thrush, Ryan Jiang, Max Bartolo, Amanpreet Singh, Adina Williams, Douwe Kiela, and Candace Ross. 2022.
\newblock Winoground: Probing vision and language models for visio-linguistic compositionality.
\newblock In \emph{Proceedings of the IEEE/CVF Conference on Computer Vision and Pattern Recognition}, pages 5238--5248.

\bibitem[{Tratz and Hovy(2010)}]{tratz2010taxonomy}
Stephen Tratz and Eduard Hovy. 2010.
\newblock A taxonomy, dataset, and classifier for automatic noun compound interpretation.
\newblock In \emph{Proceedings of the 48th Annual Meeting of the Association for Computational Linguistics}, pages 678--687.

\bibitem[{Wisniewski(1997)}]{wisniewski1997concepts}
Edward~J Wisniewski. 1997.
\newblock When concepts combine.
\newblock \emph{Psychonomic bulletin \& review}, 4:167--183.

\bibitem[{Xu et~al.(2023)Xu, Xie, Tan, Huang, Howes, Sharma, Li, Ghosh, Zettlemoyer, and Feichtenhofer}]{xu2023metaclip}
Hu~Xu, Saining Xie, Xiaoqing~Ellen Tan, Po-Yao Huang, Russell Howes, Vasu Sharma, Shang-Wen Li, Gargi Ghosh, Luke Zettlemoyer, and Christoph Feichtenhofer. 2023.
\newblock \href {http://arxiv.org/abs/2309.16671} {Demystifying clip data}.

\bibitem[{Yuksekgonul et~al.(2023)Yuksekgonul, Bianchi, Kalluri, Jurafsky, and Zou}]{yuksekgonul2023when}
Mert Yuksekgonul, Federico Bianchi, Pratyusha Kalluri, Dan Jurafsky, and James Zou. 2023.
\newblock \href {https://openreview.net/forum?id=KRLUvxh8uaX} {When and why vision-language models behave like bags-of-words, and what to do about it?}
\newblock In \emph{The Eleventh International Conference on Learning Representations}.

\bibitem[{Zhai et~al.(2023)Zhai, Mustafa, Kolesnikov, and Beyer}]{Zhai_2023_ICCV}
Xiaohua Zhai, Basil Mustafa, Alexander Kolesnikov, and Lucas Beyer. 2023.
\newblock Sigmoid loss for language image pre-training.
\newblock In \emph{Proceedings of the IEEE/CVF International Conference on Computer Vision (ICCV)}, pages 11975--11986.

\end{thebibliography}

\appendix
\section{Additional Details}
\label{sec:add-details}

\subsection{Hyper-parameter tuning for number of example captions}
\label{sec:hyper}

Table~\ref{tab:exemplar_effects} compares the results of our proposed method for a varying number of captions. While performance monotonically increases with an increasing number of diverse captions, the performance plateaus at 5 captions.

\begin{table}[H]
\centering
\resizebox{\columnwidth}{!}{%
\begin{tabular}{cccccccc}
\toprule \toprule
\textbf{\#Exemplars} & 1 & 2 & 3 & 4 & 5 & 6 & 7 \\ \midrule
\textbf{Accuracy} &    79.55 &    79.90 &    81.30 & 83.55 &    \underline{86.50} &    86.50 &    \textbf{86.55} \\ \bottomrule
\end{tabular}%
}
\caption{Effect of number of exemplars}
\label{tab:exemplar_effects}
\end{table}

\subsection{ImageNet accuracy with Retrieval with Captions}
\label{sec:imagenet}

To prove the efficacy of our proposed approach in a more general setting, we perform zero-shot classification with ImageNet with example captions for each class. Table~\ref{tab:imagenet} compares the performance of baseline template-based retrieval with CLIP with our proposed method. Our proposed method outperforms generic template-based retrieval by x\% on ImageNet.

\begin{table}[H]
\centering
\resizebox{0.75\columnwidth}{!}{%
\begin{tabular}{lc}
\toprule \toprule
Model            & Accuracy \\ \midrule
CLIP             & 71.58    \\
CLIP~\textit{w/ desc}~\cite{menon2022visual}     &  \underline{75.00}    \\
CLIP~\textit{w/ examples (ours)} & \textbf{76.85}    \\ \bottomrule
\end{tabular}%
}
\caption{Accuracy Comparison on ImageNet Dataset}
\label{tab:imagenet}
\end{table}



\subsection{Types of Compound Nouns}
\label{subsec:types}

There are three forms for compound nouns:
\begin{enumerate}
\item open or spaced - space between words (tennis shoe)
\item hyphenated - hyphen between words (six-pack)
\item closed or solid - no space or hyphen between words (bedroom)
\end{enumerate}

Table \ref{tab:compund_noun_types} shows some examples of compound nouns of different forms.

\begin{table}[h]
\centering
\resizebox{1\columnwidth}{!}{%
\begin{tabular}{@{}lllll@{}}
\toprule
\multirow{3}{*}{1. noun}       & \multirow{3}{*}{+} & \multirow{3}{*}{noun}       & bus stop        & Is this the bus stop for the number 12 bus?               \\ 
                            &                    &                             & fire-fly        & In the tropics you can see fire-flies at night.           \\ 
                            &                    &                             & football        & Shall we play footballtoday?                              \\ \midrule
\multirow{3}{*}{2. adjective}  & \multirow{3}{*}{+} & \multirow{3}{*}{noun}       & full moon       & I always feel crazy at full moon.                         \\
                            &                    &                             & blackboard      & Clean the blackboardplease.                               \\
                            &                    &                             & software        & I can't install this softwareon my PC.                    \\ \midrule
\multirow{3}{*}{3. verb(-ing)} & \multirow{3}{*}{+} & \multirow{3}{*}{noun}       & breakfast       & We always eat breakfast at 8am.                           \\
                            &                    &                             & washing machine & Put the clothes in the red washing machine.               \\
                            &                    &                             & swimming pool   & What a beautiful swimming pool!                           \\ \midrule
\multirow{3}{*}{4. noun}       & \multirow{3}{*}{+} & \multirow{3}{*}{verb(-ing)} & sunrise         & I like to get up at sunrise.                              \\
                            &                    &                             & haircut         & You need a haircut.                                       \\
                            &                    &                             & train-spotting  & His hobby is train-spotting.                              \\ \midrule
5. verb                        & +                  & preposition                 & check-out       & Please remember that check-out is at 12 noon.             \\
6. noun                        & +                  & prepositional phrase        & mother-in-law   & My mother-in-law lives with us.                           \\
7. preposition                 & +                  & noun                        & underworld      & Do you think the police accept money from the underworld? \\
8. noun                        & +                  & adjective                   & truckful        & We need 10 truckfuls of bricks.                           \\ \bottomrule
\end{tabular}}
\caption{Types of Compound Nouns}
\label{tab:compund_noun_types}
\end{table}

\subsection{List of Compound Nouns in \dataname}
\label{subsec:list}

Table~\ref{tab:compound_list} lists down all CNs in the \dataname benchmark.

\subsection{Visual examples of various categories of instances in \dataname}
\label{subsec:visual}

Fig.~\ref{fig:types_compun} illustrates 3 types of CNs used in our study: Either Noun, Both Nouns and None. A brief explanation of the 3 types is provided in Section~\ref{sec:analysis}.

\subsection{MTurk Study}
\label{subsec:mturk}

Our institution’s Institutional Review Board(IRB) has granted approval for the data collection. We will release our benchmark under the CC-BY-NC 4.0 License, which is freely available for research purposes.

{\noindent \textbf{Initial Annotator recruitment.}} We first performed a pilot run amongst 10 English-speaking MTurk annotators to test their intelligibility in identifying a set of 10 images with 10 different but commonly used compound nouns. From this study, we finally recruited 3 annotators. Our institution’s Institutional Review Board (IRB)
has approved this study. 
\vspace{1mm}

{\noindent \textbf{Removing confusing instances.}} Post collection of all instances in \dataname, the first step was to remove confusing instances where humans found it extremely difficult to distinguish between the image of a compound noun and its constituent nouns. The annotators were just asked a binary answer, i.e., confusing or not, after showing some 5 examples of confusing (e.g., cheesecake) and not confusing instances (e.g., cricket bat) to each. Finally, only instances with a majority vote of confusion amongst the 3 annotators were removed.
\vspace{1mm}

{\noindent \textbf{Human Evaluation on \dataname.}} Finally, we perform a human evaluation of our benchmark \dataname with 3 different annotators. Each annotator evaluates \dataname once, and the final reported scores in Table~\ref{tab:results} are an average of scores of all 3 annotators with the proposed evaluation metric in Equation~\ref{eqtn:eval}.


\section{Extra Details}
\label{sec:extra_details}

{\noindent \textbf{Model Parameters:}} We use CLIP-ViT-L/14 for all our experiments which have $\approx$~673M parameters with 24 and 8 encoder and decoder layers, respectively, and 16 attention heads per layer.
\vspace{1mm}

\begin{table*}[t]
\centering
\resizebox{1.9\columnwidth}{!}{%
\begin{tabular}{llllllll}
\toprule \toprule
snow ball                 & courtyard       & mountain bike    & building stone      & seat belt      & pocketknife     & teaspoon        & spray paint         \\
sun roof                  & bomb squad      & curtain rail     & bookshelf           & golf cart      & freight train   & herb scissors   & goldfish            \\
steam train               & space station   & sandpaper        & castle gate         & pasta tongs    & tailcoat        & cassette tape   & ice cap             \\
raincoat                  & thumb pin       & fruitcake        & earwig              & snow boot      & pasteboard      & shell pearl     & fur coat            \\
copper wire               & billboard       & birdhouse        & zebra crossing      & eardrum        & clotheshorse    & trash can       & Gas station         \\
firefly                   & eyeball         & streetlight      & peanut butter       & nutmeg mill    & lemon peel      & marble corridor & soup ladle          \\
windshield                & Coffee grain    & fishbowl         & chocolate crocodile & mountain goat  & watershed       & popcorn ball    & Cotton ball         \\
duckpin                   & wastebasket     & catfish          & hand brake          & sugar pea      & cement mixer    & potato salad    & floodlight          \\
pig farm                  & sand castle     & farm machine     & bullet train        & Tea cup        & Wallflower      & Ice skate       & Web page            \\
ice scoop                 & eggshell        & scoop strainer   & splatter screen     & motorcycle     & clam knife      & fishtail        & beach house         \\
blade guard               & shoe brush      & crossbow         & toothbrush          & fireman        & dogwood         & Computer mouse  & swordfish           \\
meat market               & steam iron      & football         & aircraft engine     & handbag        & pasta salad     & dirt bike       & farmhouse           \\
tennis shoes              & houseboat       & coconut haystack & tailbone            & Woodshop       & deck chair      & fingernail      & corn dog            \\
skyscraper                & metal spatula   & ice tongs        & oil pump            & saucepan       & prison dining   & water meter     & flagpole            \\
food mill                 & horsefly        & bookstore        & streetcar           & bedroom        & key chain       & pepper spray    & fishhook            \\
rubber band               & Garage door     & alphabet soup    & Bathroom sink       & Toothpaste     & egg ring        & paint brush     & corn salad          \\
sugarcane                 & headband        & lipstick         & Hairband            & Ice hockey     & silkworm        & bike rack       & clothesline         \\
garlic bread              & bow tie         & skateboard       & palm tree           & seahorse       & Candy cane      & golf ball       & cow pasture         \\
ladybug                   & snowball        & forehand         & headdress           & wiretap        & Cupboard        & dove necklace   & chocolate macaroons \\
oven mitt                 & spaceship       & toast tongs      & ballpark            & bedsheet       & pinwheel        & face mask       & pancake pen         \\
stone wall                & sunfish         & yardstick        & dishwasher          & footnote       & Snack house     & chocolate chips & earring             \\
dog house                 & shoe rack       & shellfish        & tumbleweed          & meat hammer    & snow cone       & trophy case     & dish rack           \\
panini spatula            & corner spoon    & Fish tank        & telephone cord      & ponytail       & oven tongs      & wine bar        & rolling pin         \\
rattlesnake               & gas guzzler     & almond biscotti  & honeycomb           & fingerprint    & paper clip      & Kitchen sink    & cricket bat         \\
robot arm                 & cigarette butt  & coca leaf        & oil thermometer     & coconut tree   & church bell     & cartwheel       & graveyard           \\
jet engine                & paper towel     & bankbook         & bread knife         & tablespoon     & eyelid          & Waterwheel      & toothpick           \\
dump truck                & station wagon   & hair brush       & penknife            & key card       & grasshopper     & seaweed         & banana tree         \\
greenhouse                & pasta rake      & firearm          & bus stop            & duckbill       & waterspout      & pigtail         & flower obsidian     \\
shoe box                  & shoelace        & roller coaster   & drugstore           & pumpkin gutter & gumball         & car charger     & coffee table        \\
Cotton bud                & water tank      & headline         & honeybee            & starfish       & Pool Table      & courthouse      & toadstool           \\
crab cake                 & wheelchair      & toilet paper     & fountain pen        & teapot         & moonstone       & watermelon      & caveman             \\
car tyre                  & wind turbine    & rubber spatula   & rib cage            & fire truck     & Wheelhouse      & baseball bat    & paperback           \\
piggybank                 & garbage man     & silicon chips    & jellyfish           & School bus     & relay station   & ice cream       & soapstone           \\
handcuff                  & sunflower       & wristwatch       & firewood            & banknote       & cattail         & lime juicer     & basketball court    \\
drumstick                 & alarm box       & bourbon balls    & campfire            & cowboy         & straining scoop & chain armor     & paddle wheel        \\
grill surface thermometer & spaghetti spoon & candy bar        & keyboard            & kneecap        & footprint       & car door        & waistcoat           \\
pasta scoop               & chalkboard      & candlestick      & peanut              & seafood        & skullcap        & bulldog         & Sugar plum          \\
ring finger               & jungle gym      & buttercup        & handshake           & Hair spray     & thunderhead     & Fish net        & rubber glove        \\
Ice cube                  & horseshoe       & shuttle cork     & headlamp            & headphone      & space shuttle   & cocktail spoon  & snow man            \\
street lamp               & steel drum      & Beach resort     & Ground beef         & exhaust fan    & fruit fly       & barman          & tennis court        \\
seafood scissors          & powder brush    & car factory      & manhole cover       & plastic bag    & jellybean       & backbone        & police van          \\
carpet                    & earthworm       & snowbird         & strawberry          & doorbell       & earbuds         & tar pit         & dish brush          \\
horse cart                & Cupcake         & riverbank        & ink pot             & water butt     & car phone       & pancake         & rainbow             \\
headlight                 & M \& M cookies  & bolt cutter      & eggplant            & boat house     & oyster knife    & elevator shaft  & Coffee mug          \\
railroad                  & gunpowder       & shoe shop        & wallpaper           & horse earrings & laser tag       & tapeworm        & tree house          \\
pizza wheel               & Keyhole         & Kitchen counter  & butterfly           & bullseye       & mailbox         & avocado tool    & sheep dog           \\
grapefruit knife          & necklace        & pizza lady       & letterhead          & arrowhead      & eyeglasses      & earphone        & shoe horn           \\
armchair                  & glasshouse      & fish spatula     & elephant ear        & suitcase       & exercise bike   & vanilla bean    & cassette recorder   \\
butter knife              & Table cloth     & honey dipper     & dustpan             & paper cup      & sunspot         & hornbill        & lighthouse          \\
food court                & hand grip       & fruitcup         & watercolor          & pinecone       & lab coat        & seashell        & piston ring        
 \\ \bottomrule  
\end{tabular}}
\caption{List of Compound Nouns in \dataname}
\label{tab:compound_list}
\end{table*}

{\noindent \textbf{Compute Infrastructure:}} All our experiments are conducted on a single NVIDIA A100 GPU.
\vspace{1mm}

{\noindent \textbf{Implementation Software and Packages:}} We implement all our models in PyTorch \footnote{\url{https://pytorch.org/}} and use the HuggingFace \footnote{\url{https://huggingface.co/}} implementations of CLIP. We also use the following repositories for running the baselines:
ALBEF~\cite{li2021align}\footnote{https://github.com/salesforce/ALBEF}, BLIP~\cite{li2022blip}\footnote{https://github.com/salesforce/BLIP}, CLIP~\cite{radford2019language}\footnote{https://github.com/openai/CLIP}, MetaCLIP~\cite{xu2023metaclip}\footnote{https://github.com/facebookresearch/MetaCLIP}, OpenCLIP~\cite{ilharco_gabriel_2021_5143773}\footnote{https://github.com/mlfoundations/open\_clip}, CLIP w/ Descriptors\cite{menon2022visual}\footnote{https://github.com/sachit-menon/classify\_by\_\description\_release/tree/master/descriptors}. All softwares and packages are open source and are available for academic use, and were used only for academic purpose.
\vspace{1mm}

{\noindent \textbf{Image Curation:}} We use multiple websites to curate the images for our \dataname benchmark. Some of the these websites are: \\
\hspace{2em} 1. ${https://pixabay.com}$ \\
\hspace{2em} 2. ${https://www.pinterest.com}$ \\
\hspace{2em} 3. ${https://www.wikipedia.org}$ \\
\hspace{2em} 4. ${https://www.istockphoto.com}$ \\
\hspace{2em} 5. ${https://www.britannica.com}$ \\


\vspace{1mm}

\end{document}